\documentclass[10pt,twocolumn,letterpaper]{article}

\usepackage{wacv}
\usepackage{times}
\usepackage{epsfig}
\usepackage{graphicx}
\usepackage{amsmath}
\usepackage{amssymb}

\usepackage{multicol}
\usepackage[ruled,vlined,algo2e,linesnumbered]{algorithm2e}
\SetKwComment{Comment}{$\triangleright$\ }{}
\usepackage{booktabs}
\usepackage{color,soul}
\usepackage{capt-of,etoolbox}
\usepackage{etoolbox,lipsum}

\wacvfinalcopy 

\def\wacvPaperID{381} 

\ifwacvfinal\fi
\begin{document}

\title{A Progressively-trained Scale-invariant and Boundary-aware Deep Neural Network for the Automatic 3D Segmentation of Lung Lesions}

\author{
Bo Zhou \thanks{Contributed as intern at Merck \& Co., Inc., West Point, PA, USA.}\\
School of Computer Science \\
Carnegie Mellon University\\
Pittsburgh, PA, USA\\
\and
Randolph Crawford \\
Image Data Analytics \\
Merck \& Co., Inc. \\
West Point, PA, USA \\
\and
Belma Dogdas \\
Image Data Analytics \\
Merck \& Co., Inc. \\
Rahway, NJ, USA\\
\and
Gregory Goldmacher \\
Translational Biomarkers \\
Merck \& Co., Inc. \\
West Point, PA, USA \\
\and
Antong Chen \thanks{Corresponding author: antong.chen@merck.com}\\
Image Data Analytics \\
Merck \& Co., Inc. \\
West Point, PA, USA \\
}


\maketitle
\ifwacvfinal\thispagestyle{empty}\fi

\begin{abstract}
\vspace{-0.25cm}
Volumetric segmentation of lesions on CT scans is important for many types of analysis, including lesion growth kinetic modeling in clinical trials and machine learning of radiomic features. Manual segmentation is laborious, and impractical for large-scale use. For routine clinical use, and in clinical trials that apply the Response Evaluation Criteria In Solid Tumors (RECIST), clinicians typically outline the boundaries of a lesion on a single slice to extract diameter measurements. In this work, we have collected a large-scale database, named LesionVis, with pixel-wise manual 2D lesion delineations on the RECIST-slices. To extend the 2D segmentations to 3D, we propose a volumetric progressive lesion segmentation (PLS) algorithm to automatically segment the 3D lesion volume from 2D delineations using a scale-invariant and boundary-aware deep convolutional network (SIBA-Net). The SIBA-Net copes with the size transition of a lesion when the PLS progresses from the RECIST-slice to the edge-slices, as well as when performing longitudinal assessment of lesions whose size change over multiple time points. The proposed PLS-SiBA-Net (P-SiBA) approach is assessed on the lung lesion cases from LesionVis. Our experimental results demonstrate that the P-SiBA approach achieves mean Dice similarity coefficients (DSC) of 0.81, which significantly improves 3D segmentation accuracy compared with the approaches proposed previously (highest mean DSC at 0.78 on LesionVis). In summary, by leveraging the limited 2D delineations on the RECIST-slices, P-SiBA is an effective semi-supervised approach to produce accurate lesion segmentations in 3D.
\end{abstract}

\section{Introduction}
In oncology CT and MR images are used in clinical care and clinical trials to assess the effect of treament on the size of lesions over time. In clinical trials of solid lesion indications, lesion size is assessed according to the Response Evaluation Criteria in Solid Tumors (RECIST) \cite{eisenhauer2009new}, with a single diameter measurement used to represent each “target” (quantified) lesion. For this purpose, and in clinical care settings, radiologists often manually delineate the boundary of a lesion on a single slice, where the lesion looks largest, so that the diameter measurements can be extracted from this region of interest (ROI). Nevertheless, definition of the whole lesion volume in three dimensions, a process called segmentation, is useful for a variety of applications, including lesion growth kinetic modeling and research into imaging features that may act as novel biomarkers. However, delineating lesions in 3D medical images is a labor- and resource-intensive task, which prohibits the volumetric assessments from replacing the RECIST-based assessments. Although many works have been proposed to produce lesion segmentations in CT images using region-based, graph-based, or deformable model-based approaches \cite{hoogi2017levelset, nithila2016fuzzy, cha2018graphcuts, gu2013automated}, the performances of these conventional approaches are highly sensitive to the initial conditions and parameter settings of models. Therefore, it is highly desirable to develop an automatic 3D lesion segmentation algorithm that can provide accurate, robust, and consistent results.

\begin{figure}[!htb]
\centering
\includegraphics[width=0.47\textwidth]{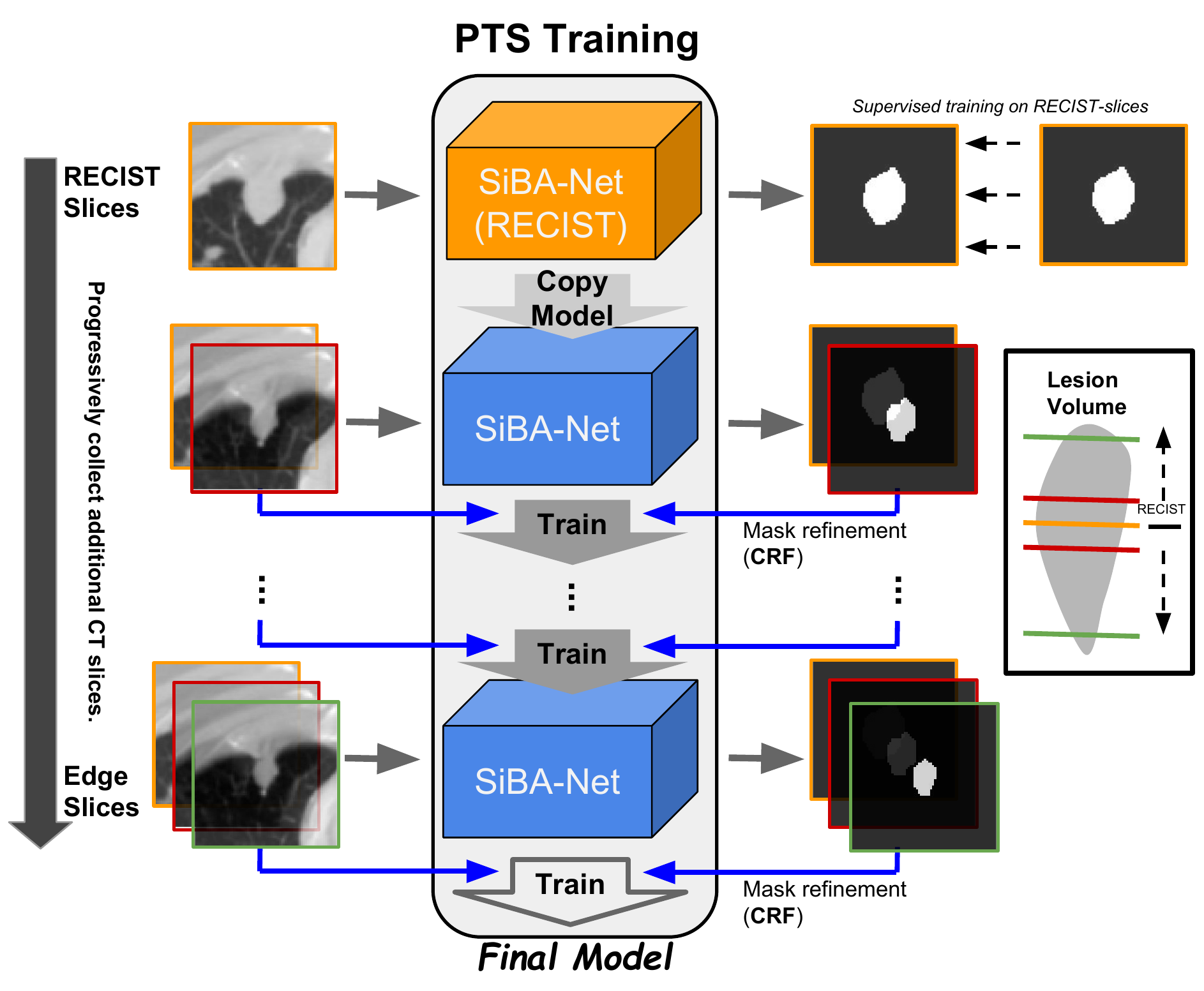}
\caption{: Overall pipeline of the proposed progressive lesion segmentation (PLS) approach using a scale-invariant and boundary-aware network (SiBA-Net). The initial SiBA-Net (orange block) is first trained on RECIST-slices (orange boarder/line) with ground-truth 2D lesion delineations. Then after convergence an instance of the SiBA-Net (blue block) is copied for predicting the segmentation on the neighboring slices. The segmentations are refinement by CRF (red boarder/line) and added to the training dataset with the corresponding images for the next iteration of training. The procedure repeats until no additional training samples are added, and the final model is obtained upon convergence.}
\label{fig_intro}
\end{figure}

In this work, we aim to develop an automatic volumetric lesion segmentation algorithm by leveraging the 2D ROIs created by radiologists during RECIST evaluation of clinical CT images. First of all, we propose a progressive lesion segmentation (PLS) algorithm that iteratively generates 2D lesion segmentations from the RECIST slice to the neighboring slices, and eventually to the edge slices in a semi-supervised manner. Secondly, we introduce a scale-invariant and boundary-aware network (SiBA-Net) that can effectively cope with the scale-variability of lesion when the PLS propagates from the RECIST slices to the edge slices or is applied to segment lesions with a large range of size variability, e.g. lesions demonstrating positive response to treatments and scanned longitudinally. The overall workflow of the algorithm is shown in Figure \ref{fig_intro}. Specifically, the SiBA-Net is trained starting with the 2D delineations on the RECIST-slices, and the trained network is applied on the neighboring CT slices to make predictions. Then the predictions are refined using fully-connected conditional random field (FC-CRF) \cite{krahenbuhl2011efficient}, and incorporated with the corresponding image slices into the training data to updated the SiBA-Net. The training-predicting-updating steps are executed iteratively until no additional training samples are added. Although the sections of the lesion shrinks when the PLS progresses from the RECIST slice to the superior and inferior edges of the lesion in 3D, the scale-invariance mechanism ensures that the convolution filters at various scales will be able to generate response to the lesion region, which brings robustness against lesion size variability to the PLS. In short, the proposed PLS algorithm based on SiBA-Net, namely P-SiBA, is a self-taught and robust 2D-to-3D segmentation approach that can effectively leverage the 2D segmentations on the RECIST-slices to automatically generate the whole volumetric segmentations for lesions.

To develop and evaluate the proposed P-SiBA approach, we have collected a large set of clinical CT scans with 2D delineations on the RECIST-slices. The dataset, named LesionVis, consists of manual 2D delineations on the RECIST slice for 14,961 lung, liver, mediastinum, subcutaneous, and abdomen lesions, as well as enlarged lymph nodes. Specifically, we identify 4,001 lung lesions for training the P-SiBA algorithm. To further assess the proposed approach on LesionVis, we conduct 3D manual delineation for 250 lung lesions in this dataset. Our results demonstrate that the proposed P-SiBA approach is able to achieve mean Dice similarity coefficients (DSC) of 0.81 on the LesionVis dataset, which significantly outperforms the previously published methods \cite{cai2018accurate}.

\subsection{Contributions}
\noindent The major contributions of this work include:
\begin{itemize}
\item We develop an automatic volumetric lesion segmentation algorithm by leveraging 2D delineations on the RECIST-slices as training data.

\item We propose a progressively-trained end-to-end scale-invariant and boundary-aware network that can effectively cope with the scale variance of a lesion when the segmentation propagates from RECIST slices to edge slices. The SiBA-Net is also effective when it is applied to segment lesions with a large range of size variability, e.g. lesions that demonstrate positive response to treatments and scanned longitudinally.

\item We collect a large-scale database with manual 2D lesion delineations on the RECIST-slices, which is named LesionVis. An extensive experimental analysis of our approach is performed on this dataset, and it is shown by both qualitative and quantitative evaluations that the proposed approach is able to produce automatic volumetric lesion segmentations that are more accurate than the current state-of-the-art approaches.
\end{itemize}

\section{Related Works}
\textbf{Automatic lesion segmentation in CT} is one of the most important topics in medical image analysis. Tremendous efforts have been made on the development of lesion classification and segmentation methods in different organs, such as brain \cite{shen2017boundary,kamnitsas2017efficient}, breast \cite{yuan2007dual,horsch2001automatic}, lung \cite{gu2013automated,armato2004automated}, and liver \cite{li2013likelihood}. Recently, \cite{yan2018deep} presented a large-scale lesion dataset, called DeepLesion, which is similar to our LesionVis dataset but doesn't include pixel-wise 2D delineations on the RECIST-slices. Based on the DeepLesion dataset, \cite{cai2018accurate} proposed a weakly supervised self-paced segmentation segmentation (WSSS) method that utilized tumor's long- and short-axis drawings on the RECIST-slices to generate the full 3D lesion segmentations. In the first step, WSSS produces the 2D lesion segmentations on the RECIST-slices using unsupervised methods, such as GrabCut \cite{rother2004grabcut}. Then, given the initial 2D segmentations on the RECIST-slices, a CNN \cite{nogues2016automatic} adopted from \cite{xie2017holistically} is trained for predicting lesion segmentations of neighboring CT slices in an iterative manner. Validated on a relatively small dataset (200 lesions segmented in 3D), WSSS was shown to be able to generate 3D lesion segmentations. However, inaccurate and non-robust initial segmentations on the RECIST-slices could potentially undermine the approach from the first step which could lead to deteriorated performance. In addition, the conventional CNN architecture used in WSSS, such as \cite{nogues2016automatic}, doesn't consider either the 2D lesion scale variation from RECIST slices to edge slices, or 3D lesion size variabilities especially at different time points of treatments. In comparison, our collected LesionVis dataset with manual 2D lesion delineations on the RECIST-slices can avoid the issue of failures to segment the lesions on the RECIST slices. For dealing with the scale-variation issue, the scale-invariant CNN architecture design could play an important role. \cite{xu2014scale,kanazawa2014locally} proposed the earliest scale-invariant CNN with the key concept of either transforming the filters in convolutional layers or transforming the CNN's perception field to generate the same network response when the object's scale varies. Our proposed SiBA-Net adopts the scale-invariant architecture with multi-column paths from \cite{xu2014scale}.

\textbf{Semi/Weakly-supervised learning} has important applications in the detection and segmentation of objects, especially when the amount of labeled data is limited. Semi-supervised learning algorithms \cite{bai2017semi,sedai2017semi,zhou2018weakly,oquab2015object} for the segmentation of regions with disease, such as Cardiac-FCN\cite{bai2017semi} and SG-VAE\cite{sedai2017semi}, rely on special training strategies to effectively leverage the labeled data and extend to unlabeled data during the training. For instance, a CNN can be trained on existing annotated training data to capture the corresponding data distribution. The converged CNN is then iteratively updated and enhanced using the unlabeled data. Recently, using this strategy, \cite{bai2017semi} presented a novel method to improve the 2D segmentation with additional training data generated from unlabeled cardiac MR images with conditional random field. Their results showed significant improvement on 2D segmentation accuracy when the semi-supervised strategy was incorporated. \cite{sedai2017semi} proposed another type of semi-supervised learning approach by minimizing the difference between the distribution of training data and the distribution of unlabeled data in their latent space using a variational autoencoder (VAE). However, unlike the 2D optic cup segmentation task addressed in \cite{sedai2017semi}, 2D sections on a lesion can exhibit drastic morphological changes across CT slices, which may not be modeled properly by the existing training data from the RECIST-slices. Therefore, we adopt the merit of expandability from \cite{bai2017semi} by employing a semi-supervised progressive learning scheme to gradually achieve the lesion segmentations from RECIST-slices to edge slices.

\section{Method}
We introduce a novel algorithm, named P-SiBA, to learn a CNN model for performing volumetric lesion segmentation using 2D RECIST-slices with pixel-wise manual annotations. Leveraging the delineations on the RECIST-slices, the approach is able to draw additional training data from the remaining unlabeled 3D volume to help improve the performance of the model. Our proposed P-SiBA algorithm consists of three major steps: \textbf{(i)} Training of a SiBA-Net based on the 2D delineations on the axial RECIST-slices. \textbf{(ii)} Progressively expanding the training dataset by adding neighboring CT slices and predictions on them that are refined by CRF. \textbf{(iii)} Post-processing to refine the segmentation results in 3D. The general pipeline of P-SiBA is shown in Figure \ref{fig_intro}, with details described in the following sections.

\begin{figure*}[!htb]
\centering
\includegraphics[width=0.83\textwidth]{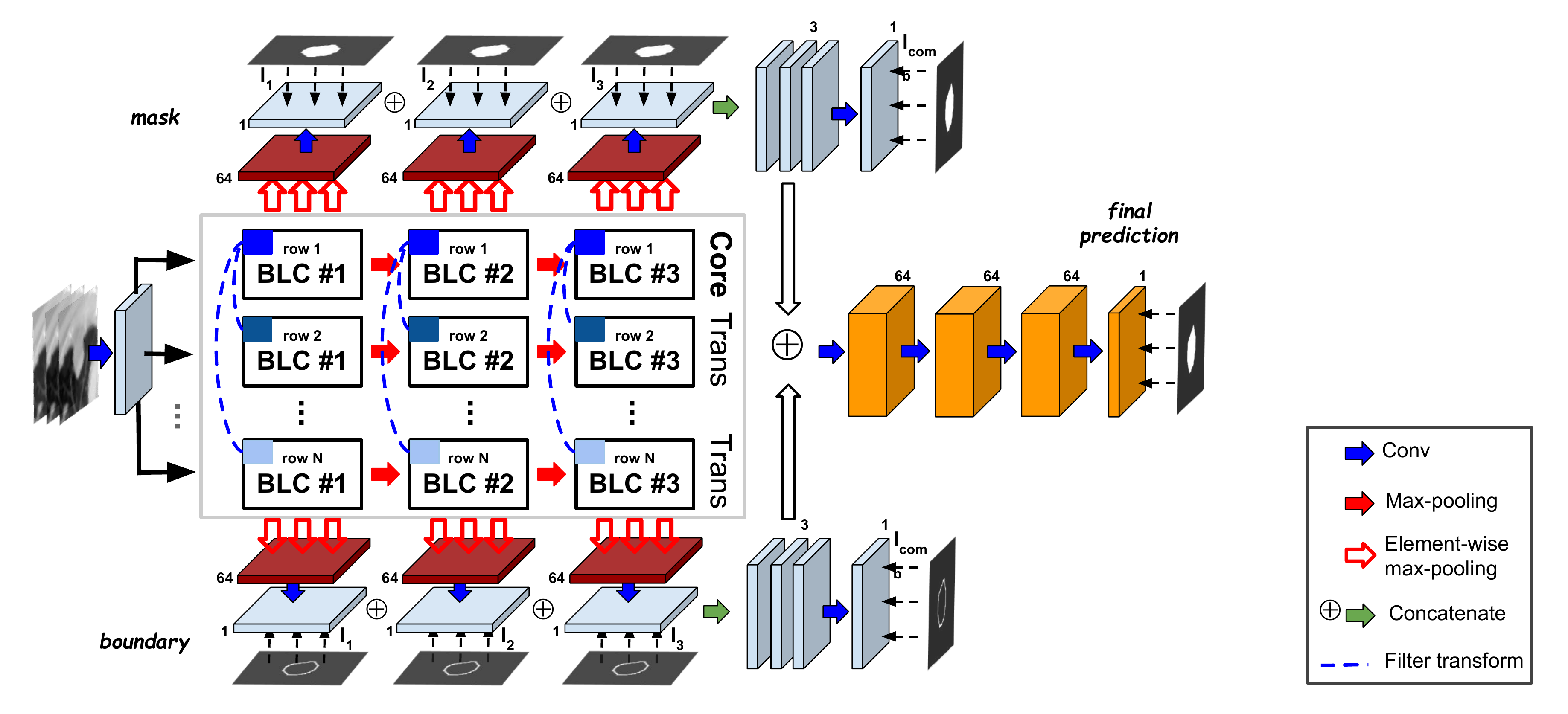}
\caption{The architecture of the proposed SiBA-Net. The response after an initial convolutional layer is passed into multiple branches in VGG16 architecture (details shown in Figure \ref{fig_VGG}) with three different levels of outputs from each branch. Same-level outputs from the branches are combined by element-wise max-pooling to generate the scale-invariant responses. The outputs at the top and bottom are deeply supervised by the lesion’s segmentation mask and boundaries derived from the mask, respectively. Meanwhile, the intermediate features from the two deep supervision networks from top and bottom are combined to generate a final lesion segmentation mask.}
\label{fig_SiBA}
\end{figure*}

\begin{figure}[!htb]
\centering
\includegraphics[width=0.46\textwidth]{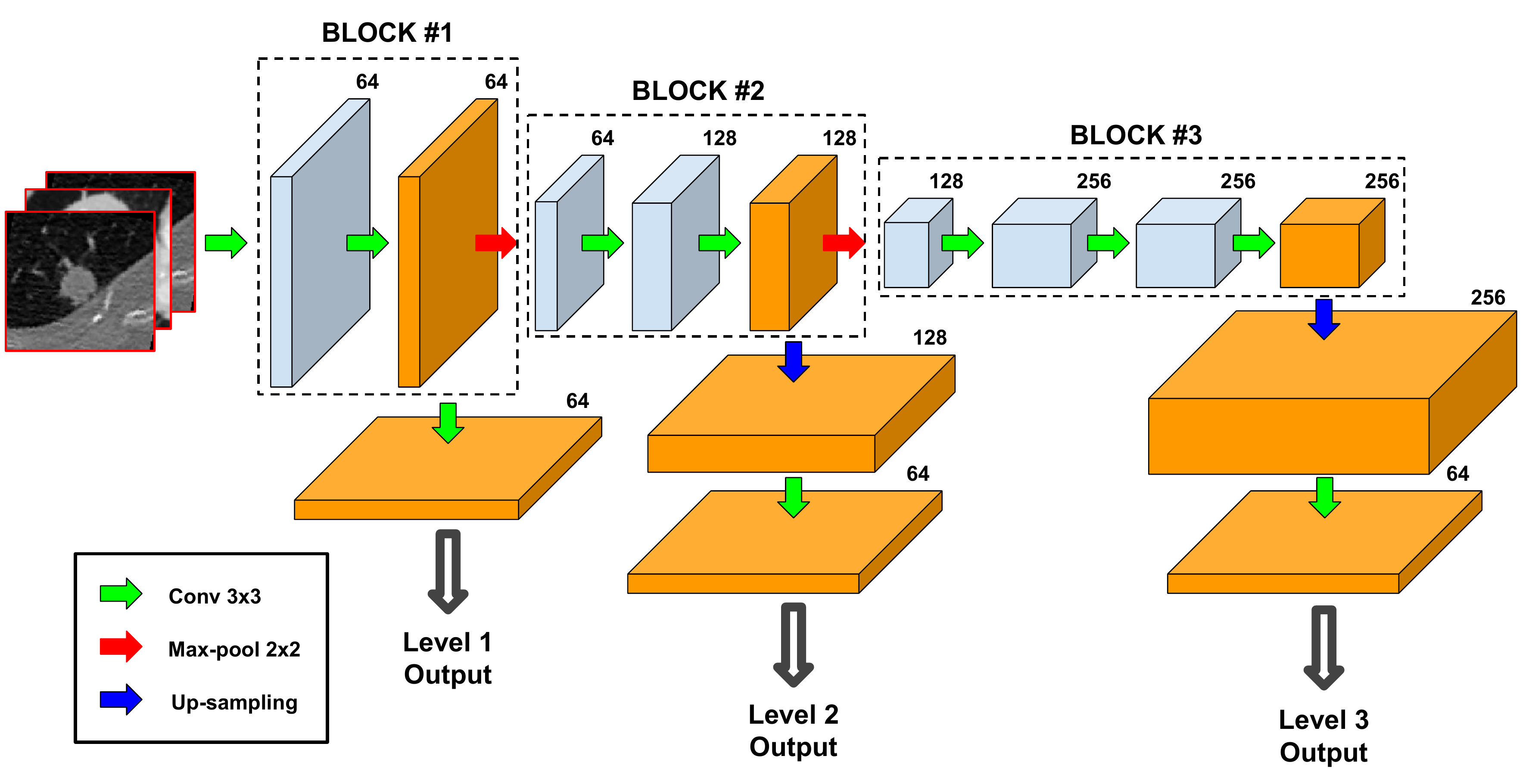}
\caption{The adopted VGG16 architecture used in the proposed SiBA-Net. Three levels of outputs are generated from the three blocks of VGG16. The outputs are convolved and re-sampled to the same size, such that element-wise max-pooling can be performed in the following steps.}
\label{fig_VGG}
\end{figure}

\begin{figure*}[!htb]
\centering
\includegraphics[width=0.9\textwidth]{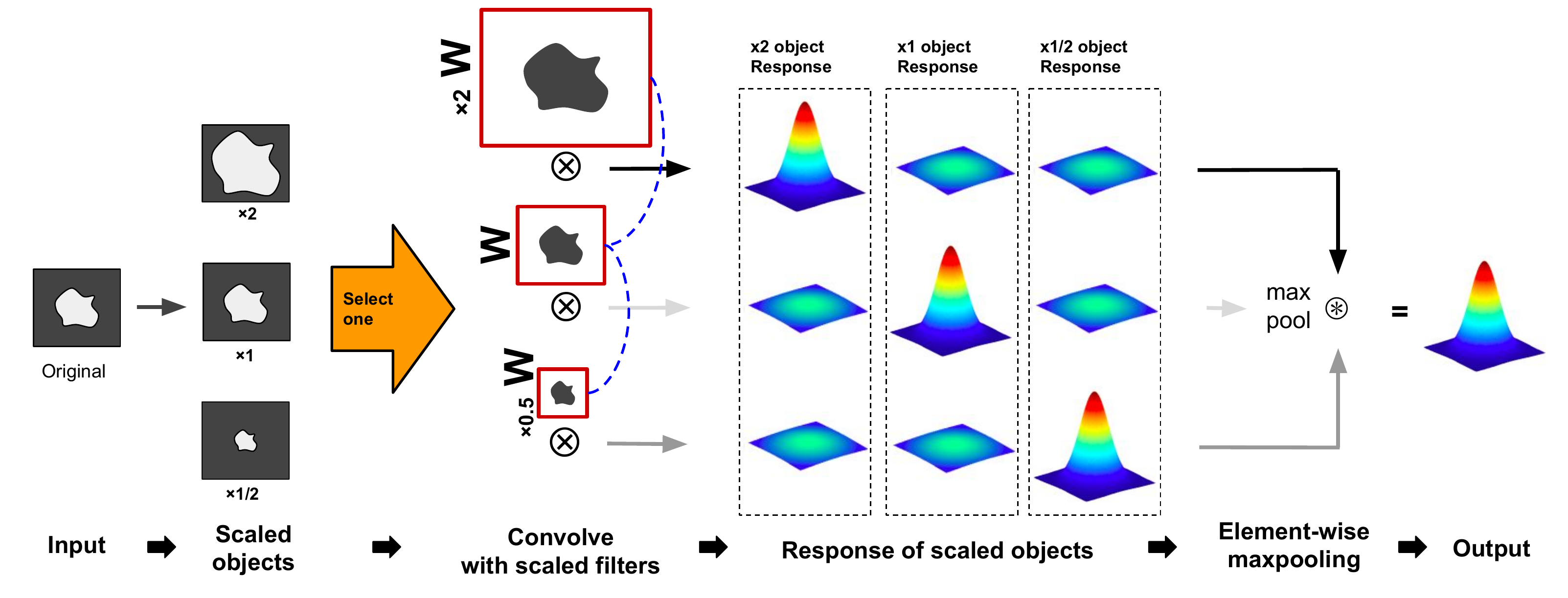}
\caption{An illustration of how the scale-invariant network architecture can robustly respond to patterns of tumor at various scales: In a standard CNN, filter $W$ will respond to the $\times 1$ lesion because of the scale similarity between $W$ and the $\times 1$ lesion. However, in our scale-invariant architecture, the filters transformed/scaled from $W$ can respond well to the $\times \frac{1}{2}$ and $\times 2$ lesions as well. With the element-wise max-pooling operation between multiple scales at the end, the final output response is independent of lesion scale, and therefore the SiBA-Net is able to robustly respond to lesion sections with varying sizes on the successive CT slices.}
\label{fig_scaleinv}
\end{figure*}

\subsection{Scale-invariant and boundary-aware CNN}
In conventional CNN structures, dealing with objects at multiple scales is achieved jointly by the pooling layers and the convolutional layers. The convolutional layers not only need to learn different features of target objects but also their scale variations when objects at different scales are present, which could lead to large models that are challenging to train. One of the most common strategies to deal with the scale variation problem is scale-jittering. \cite{krizhevsky2012imagenet}. Inspired by the invariance-to-shift property of the convolutional operation, we propose to compute convolutions on the images with filters at multiple scales. This is done by adding additional branches, each of which a conventional CNN with filters at a different scale. The filter weights on the branches are regularized and tied to a primary branch, and essentially the amount of trainable parameters remain the same as the architecture with single branch. As such, the scale-invariance property is brought into our model, requiring neither additional data augmentation nor increased model size. 

Our proposed SiBA-Net uses 3 branches of CNNs with various filter sizes to capture lesions at various scales in the training data. The architecture of the SiBA-Net is illustrated in Figure \ref{fig_SiBA}. From left to right, the input image is fed into all CNN branches. To avoid overfitting, we adopted a VGG16 model pre-trained on the ImageNet dataset \cite{simonyan2014very} for each branch with multi-level outputs as shown in Figure \ref{fig_VGG}. The core-branch adopts $3\times 3$ filters in all the convolutional layers, and the other two branches adopt convolutional filters scaled up from the corresponding filters on the core-branch to the sizes of $5\times 5$ and $7\times 7$ using the filter transformation method in \cite{xu2014scale}. As shown in Figure \ref{fig_scaleinv}, the outputs of the 3 CNN branches are processed by element-wise max-pooling to generate the eventual scale-invariant response, and all 3 branches essentially share a common set of weights in the convolution filters. As a result, the patterns of a single lesion image should be captured by one or more branches in the network, which brings robustness to the segmentation of lesion images at different scales.

The training of the SiBA-Net is jointly optimized and deeply supervised by the lesions' segmentation masks and the boundaries derived from the masks. Therefore the joint loss function is defined in Equation \ref{eq:loss}:
\begin{equation}
\label{eq:loss}
\begin{aligned}
\boldsymbol{\mathcal{L}} = w_{m} \sum_{n \in l_1,l_2,l_3,l_{com}} (1 - \frac{2|M_n \cap \hat{M_n}|}{|M_n| + |\hat{M_n}|}) \\
+ w_{b} \sum_{n \in l_1,l_2,l_3,l_{com}} (1 - \frac{2|B_n \cap \hat{B_n}|}{|B_n| + |\hat{B_n}|}) \\
+ w_{f} (1 - \frac{2|L \cap \hat{L}|}{|L| + |\hat{L}|})
\end{aligned}
\end{equation}

\noindent where $\hat{M_n}$ and $\hat{B_n}$ represent the regional and boundary predictions, respectively. As it is shown in Figure \ref{fig_SiBA}, these predictions are generated with outputs from the three levels of deep supervisions, namely $l_1$, $l_2$, and $l_3$, as well as from their combination, namely $l_{com}$. In addition, $\hat{L}$ is the final segmentation prediction made from the combination of all features. Terms $M_n$, $B_n$, and $L$ represent the corresponding ground-truths. Terms $w_m$, $w_b$, $w_f$ are the weights determining the contributions of each type of loss in the final loss function, and we set $w_m = w_b = w_f = 1$ in our training. 

\subsection{Progressive lesion segmentation (PLS)}

\vspace{0.2cm}
\noindent \textbf{Lesion axial range estimation:} With the knowledge that RECIST-slice is selected on the cross section showing the longest diameter of the lesion, and since the lesions generally have a spheroid or ellipsoid shape in 3D, we approximate the range for evaluating successive CT slices on the axial direction by calculating the maximum diameter of the 2D delineation on the RECIST-slice. Based on the length of the diameter $d$ (converted into mm), we define the axial range for the evaluation to be $[-0.8d, 0.8d]$ from the RECIST slice on the axial direction. Normalized by the slice thickness, all 2D CT slices falling into this range are included and evaluated in the volumetric lesion segmentation task.

\vspace{0.2cm}
\noindent \textbf{Progressive segmentation training:} In order to obtain the 3D lesion segmentations from the 2D delineations on RECIST-slices, we train our SiBA-Net using a self-paced algorithm. The progressively-trained algorithm P-SiBA begins by training a SiBA-Net with 2D lesion delineations on the RECIST-slices. After the training for the initial SiBA-Net converges, we use the model to predict 2D segmentations on the successive CT slices. Then for each segmentation we apply a fully-connected conditional random field (FC-CRF) using both the current prediction and the corresponding image as inputs to generate a refined segmentation with tight boundary. The refined segmentations and the corresponding images are then added to the training set to update the SiBA-Net in the next iteration. 

We summarize our P-SiBA algorithm in Algorithm \ref{alg:train_alg}: A SiBA-Net is first trained on the set of RECIST-slices $T = \{(I^R_i, L^R_i)\}$. Then, the converged model generates the lesions' segmentations in the form of probability maps $L_i^{\alpha+R}$ on the neighboring CT slices. Given both the probability maps $L_i^{\alpha+R}$ and the images $I_i^{\alpha+R}$, the FC-CRF is applied to refine the 2D segmentations, and the segmentations and the images are incorporated into the training set $T \cup \{(I^{\alpha+R}, L^{\alpha+R})$. The procedure iterates as $\alpha$ increases from $0$ to the slice index at the end of axial range. Eventually we are able to obtain the converged model on all training images gradually.

\begin{algorithm2e*}[!htb]
\caption{Progressive Segmentation Training.}\label{alg:train_alg} 

\textbf{Input:} $\Pi$ = \{($V_i$, $L^R_i$)\}, for $i \in \{1,\ldots,N\}$ \Comment*[r]{ Training volumes and 2D RECIST delineations}

\textbf{Initialize \#1:} {$D_i \gets$ $f_{long}(V_i, L^R_i)$}, for $i \in \{1,\ldots,N\}$  \Comment*[r]{Axial range estimation}

\textbf{Initialize \#2:} (i) Maximal \# of progressive iteration: $K \gets D$; (ii) load SiBA-Net with pre-trained parameters: $\theta_0$

$T = \{(I^R_i, L^R_i)\} \gets \Pi$  \Comment*[r]{RECIST-slice training set extraction}

$\theta_{init} :\gets min [L(T, \theta_0)]$  \Comment*[r]{Optimize loss until converge on RECIST-slices}

\For{$k = 1$ to $K$}
{
	\For{$\alpha = -k$ to $k$}
    {
     $L_i^{\alpha+R} \gets f(I_i^{\alpha+R}, \theta_{k-1})$  \Comment*[r]{CNN inference on neighboring slices}
     $L_i^{\alpha+R} \gets CRF(L_i^{\alpha+R}, I_i^{\alpha+R})$  \Comment*[r]{Segmentation optimization}
     $T \gets T \cup \{(I^{\alpha+R}, L^{\alpha+R})\}$  \Comment*[r]{Add new data into training set}
    }
$\theta_{k} :\gets min [L(T, \theta_{k-1})]$  \Comment*[r]{Optimize loss until converge on updated training set}
}

\textbf{Output} $\theta_{K}$  \Comment*[r]{Return final SiBA-Net upon convergence}

\end{algorithm2e*} 

\subsection{Post-processing and segmentation refinement}
After obtaining the volumetric segmentations from P-SiBA, we add a post-processing step for eliminating the potential minor irregularities between slices in the axial direction as a byproduct of the slice-by-slice prediction mechanism. We use two criteria to determine if a new 2D segmentation along the axial direction is valid: Progressively moving from the delineations on the RECIST-slices to the successive CT slices at the anterior and inferior edges of the lesion, we keep a new 2D segmentation if and only if (i) the segmentation mask or the mask's centroid lies inside the previous slice's segmentation mask; and (ii) the area ratio of between the new and the previous 2D segmentations is in the range of $[0.7, 1.3]$. Otherwise, we copy the previous 2D segmentation onto the new image slice as an initialization for the CRF-based refinement to obtain the new segmentation.

\section{Experiments and Results}
\subsection{Data and Training Details}
\noindent \textbf{Dataset:} We collected a large-scale LesionVis dataset in-house, which consists of 6,049 CT studies on 1,350 patients. The CT scan acquisitions were performed according to parameters that were provided to clinical trial sites, with real-time quality checks by a commercial imaging core laboratory, and typical for clinical standard-of-care imaging. Scans covered the chest, abdomen, and pelvis with slice thickness of 5 mm or less. Intravenous contrast was used except when medically contraindicated. Manual delineations on the RECIST slices are available for 14,961 lung, liver, mediastinum, subcutaneous, and abdomen lesions, as well as enlarged lymph nodes. For training and evaluating the proposed P-SiBA approach, a set of 4,001 lung lesions were identified to construct a training dataset. From the training set, a subset of 250 lung lesions (about 4,300 2D axial CT slices in total) were selected randomly and delineated manually in a slice-by-slice manner to form a dataset of 3D delineations for validating the proposed P-SiBA approach in 3D.

\vspace{0.2cm}
\noindent \textbf{Pre-processing:} For the LesionVis dataset, given each 2D delineation on the RECIST-slice, we cropped the region of interest (ROI) on the image to a square bounding box with each edge at two times the longest diameter of the lesion, such that sufficient information in the image was preserved.

The intensity of each CT scan was first converted into Hounsfield unit (HU) using the linear transformation parameters recorded in the DICOM header. Then we shifted the dynamic range by $+1,000$ and cut the intensity off at 0 to make all values non-negative. We further normalized the data by dividing the intensity values by 3,000 and cut the maximum off at 1, such that all the intensity values were normalized to the range of $[0, 1]$ while the relative intensity changes between voxels were preserved.

\vspace{0.2cm}
\noindent \textbf{Training details:} The P-SiBA algorithm was implemented in PyTorch (\textit{https://pytorch.org/}). The weights of SiBA-Net were initialized from a VGG16 network pre-trained on the ImageNet dataset to avoid over-fitting. We used stochastic gradient descent (SGD) optimizer for training the SiBA-Net with an initial learning rate of $2 \times 10^{-4}$ which was reduced at a factor of $\frac{1}{2}$ every 1,000 epochs. A batch size of 48 was used. All trainings were conducted on a single NVIDIA Pascal Quadro P6000 GPU with 24 GB memory.

\subsection{Performance Evaluation}
For quantitative evaluation, we measure the Dice similarity coefficient (DSC) and volumetric similarity (VS) to assess our algorithm's segmentation performance. DSC and VS are defined in Equation \ref{eq:dsc} and Equation \ref{eq:vs}, respectively:

\begin{equation}
\label{eq:dsc}
\begin{aligned}
DSC = \frac{2|A \cap B|}{|A| + |B|} = \frac{2 \times TP}{2 \times TP + FN + FP}
\end{aligned}
\end{equation}

\begin{equation}
\label{eq:vs}
\begin{aligned}
VS = 1- \frac{|FN - FP|}{2 \times TP + FN + FP}
\end{aligned}
\end{equation}

\noindent where A, B are the predicted segmentation and the ground-truth segmentation, respectively. Moreover, we evaluate the quality of automatic segmentations at the  boundary using Hausdorff Distance (HD), which measures the largest Euclidean distance between A and B. 

For comparison purposes, two mainstream CNN models were selected as baseline solutions: 1) U-Net \cite{ronneberger2015u}, a classical CNN for biomedical image segmentation; and 2) Holistically-nested Network (HNN) \cite{xie2017holistically}, a classical network specializing in edge detection and learning multi-scale and multi-level image feature. We also compared our algorithm with the WSSS \cite{cai2018accurate} algorithm, which implemented a double-HNN structure as their core CNN.

\vspace{0.2cm}
\noindent \textbf{Segmentations on the RECIST-slices:} Although the 2D manual delineations on the RECIST-slice are already provided in the LesionVis dataset, we would like to make sure that the proposed SiBA-Net is capable of generating accurate 2D segmentations on the RECIST-slices. The mean and standard deviation of accuracy measurements on the RECIST-slices of the LesionVis validation set (250 lesions) are shown in Table 1. It is seen that SiBA-Net outperformed both U-Net and HNN by a large margin. Then we assessed the effect of using CRF to refine the segmentation results from SiBA-Net. Specifically, we used the original prediction from SiBA-Net as the unary potential, and used the voxel values from the CT image to compute the pairwise potential, and then combined the two potentials to generate the final segmentation. As we can see from Table 1, combining SiBA-Net and CRF significantly improved the DSC values compared with results obtained using SiBA-Net alone. Therefore, the SiBA-Net + CRF approach was shown to be effective in producing accurate 2D segmentations on the RECIST-slices without any additional human interaction other than the selection of RECIST-slices and the approximate area for the lesions which were all included in the original 2D RECIST delineation step. Notice that WSSS was not compared here since its approach for segmenting the lesions on the 2D RECIST slices depended on the drawing of the long- and short-axises, which was not provided in the LesionVis dataset.

\begin{table} [!htb]
\centering
\caption{DSC on 2D RECIST-slices of LesionVis validation dataset.}
\label{tab:t_recist}
\setlength{\tabcolsep}{6pt}

\begin{tabular}{l c c c c c}
\hline
\rule{0pt}{1.1\normalbaselineskip}
\textbf{Method}             & Mean $\pm$ Std\\
\hline
\rule{0pt}{1.1\normalbaselineskip}
U-Net \cite{ronneberger2015u}     & $0.838 \pm 0.09$ \\
\rule{0pt}{1.1\normalbaselineskip}
HNN \cite{xie2017holistically}      & $0.841 \pm 0.10$ \\
\rule{0pt}{1.1\normalbaselineskip}
SiBA-Net    & $0.872 \pm 0.08$ \\
\rule{0pt}{1.1\normalbaselineskip}
SiBA-Net + CRF    & $\mathbf{0.893 \pm 0.06}$ \\ [0.1cm]
\hline

\end{tabular}
\end{table}

\begin{figure}[!htb]
\centering
\includegraphics[width=0.5\textwidth]{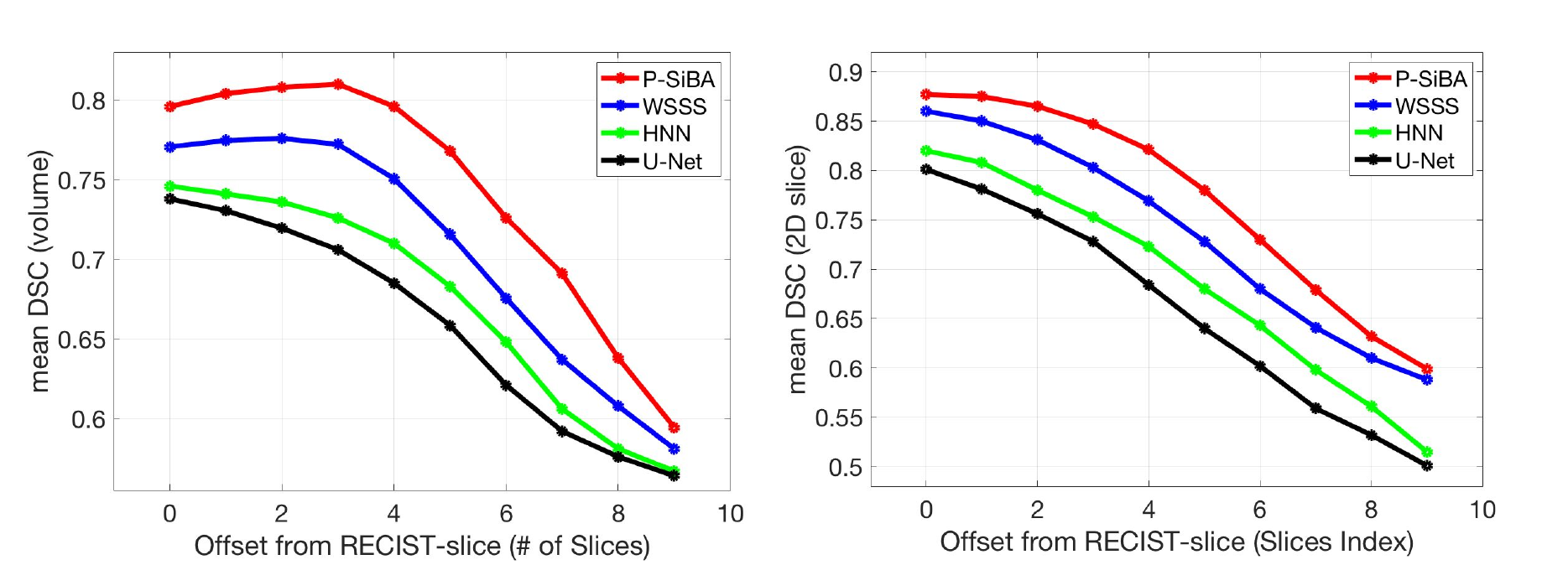}
\caption{Progressive segmentation training with different numbers of successive CT slices at different absolute offsets to the RECIST slices. Left: volumetric segmentation performance when various numbers of successive slices were included. Right: 2D segmentation performance on successive slices over progressive training process.}
\label{fig_segvol}
\end{figure}

\vspace{0.2cm}
\noindent \textbf{Progressive lesion segmentation results:} The performance of the PLS approach as described in Algorithm \ref{alg:train_alg} is evaluated quantitatively on the validation dataset. Firstly, we evaluated the volumetric lesion segmentation accuracy when the models were trained by adding successive CT slices within a range defined by the absolute slice number offset to the RECIST slices. Notice that only the successive slices within the estimated lesion axial range could be considered as potential training data for the progressive training approach. From Figure \ref{fig_segvol} it is observed that the progressive training substantially improved the volumetric segmentation accuracy of SiBA-Net. In comparison, we also implemented three other approaches, U-Net, HNN, and WSSS, and trained the networks in a progressive manner, while all three approaches were substantially outperformed by P-SiBA. It is also shown that the performance of PLS peaked when up to 3 successive slices on each side of the RECIST slices were added to the training set. 

We also evaluated the segmentation performance on successive slices over the progressive training. As shown in Figure \ref{fig_segvol}, SiBA-Net was able to robustly produce more accurate segmentations compared with other approaches. However, all four approaches had difficulties generating accurate results when distant slices were incorporated. This could be due to the deterioration of segmentation quality at the superior and inferior edges of lesions where the size and appearance properties in the CT slices significantly deviated from the RECIST slices.

\begin{figure*}[!htb]
\centering
\includegraphics[width=1\textwidth]{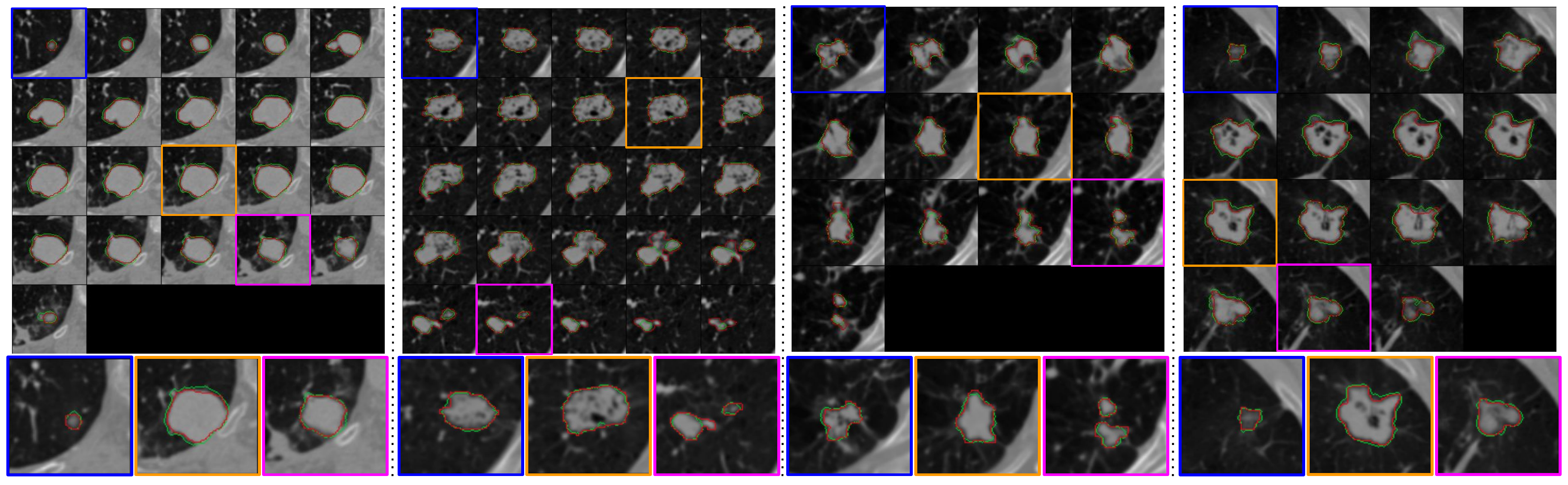}
\caption{Visualization of lesion segmentations obtained using P-SiBA (green contours) compared to the ground-truth segmentations (red contours). 3D CT volumes with segmentation contours are displayed in 2D axial view. Sample slices ranging from slices that are close to the RECIST slices (orange) and slices that are away from the RECIST slices (pink, blue) are amplified. It is shown that P-SiBA can produce accurate and robust segmentation results.}
\label{fig_VisSeg}
\end{figure*}

As a result of the progressive segmentation training, the best 3D segmentation results by all 4 approaches, U-Net, HNN, WSSS, and P-SiBA, are recorded and reported in Table \ref{tab:t_volseg} using DSC, HD, and VS as measurements of accuracy. It is seen that P-SiBA significantly outperformed the other 3 approaches when measured by all 3 categories. P-SiBA achieved mean DSC of $0.809$ and mean VS of $0.883$ on the LesionVis validation set, demonstrating its ability to produce accurate segmentations at the volumetric level. The mean HD at 7.612 mm is also the lowest among all approaches, indicating its robustness against boundary-level irregularities. Example visualizations of P-SiBA segmentation results are shown in Figure \ref{fig_VisSeg}. 

\begin{table*} [!htb]
\centering
\caption{Comparison of 3D tumor segmentation performance on the LesionVis validation dataset.}
\label{tab:t_volseg}
\setlength{\tabcolsep}{7pt}

\begin{tabular}{l c c c c c c c}
\hline
\rule{0pt}{1.1\normalbaselineskip}
\textbf{Method}                         & DSC & HD (mm) & VS \\ [0.1cm]
\hline
\rule{0pt}{1.1\normalbaselineskip}
U-Net \cite{ronneberger2015u}     & $0.745 \pm 0.10$ & $8.831 \pm 8.46$ & $0.790 \pm 0.15$ \\
\rule{0pt}{1.1\normalbaselineskip}
HNN \cite{xie2017holistically}      & $0.748 \pm 0.09$ & $8.962 \pm 8.84$ & $0.794 \pm 0.13$ \\
\rule{0pt}{1.1\normalbaselineskip}
WSSS \cite{cai2018accurate}     & $0.776 \pm 0.11$ & $8.116 \pm 6.72$ & $0.838 \pm 0.12$ \\
\rule{0pt}{1.1\normalbaselineskip}
P-SiBA    & $\mathbf{0.809 \pm 0.12}$ & $\mathbf{7.612 \pm 5.03}$ & $\mathbf{0.883 \pm 0.13}$ \\ [0.1cm]
\hline

\end{tabular}
\end{table*}

\begin{figure}[!htb]
\centering
\includegraphics[width=0.48\textwidth]{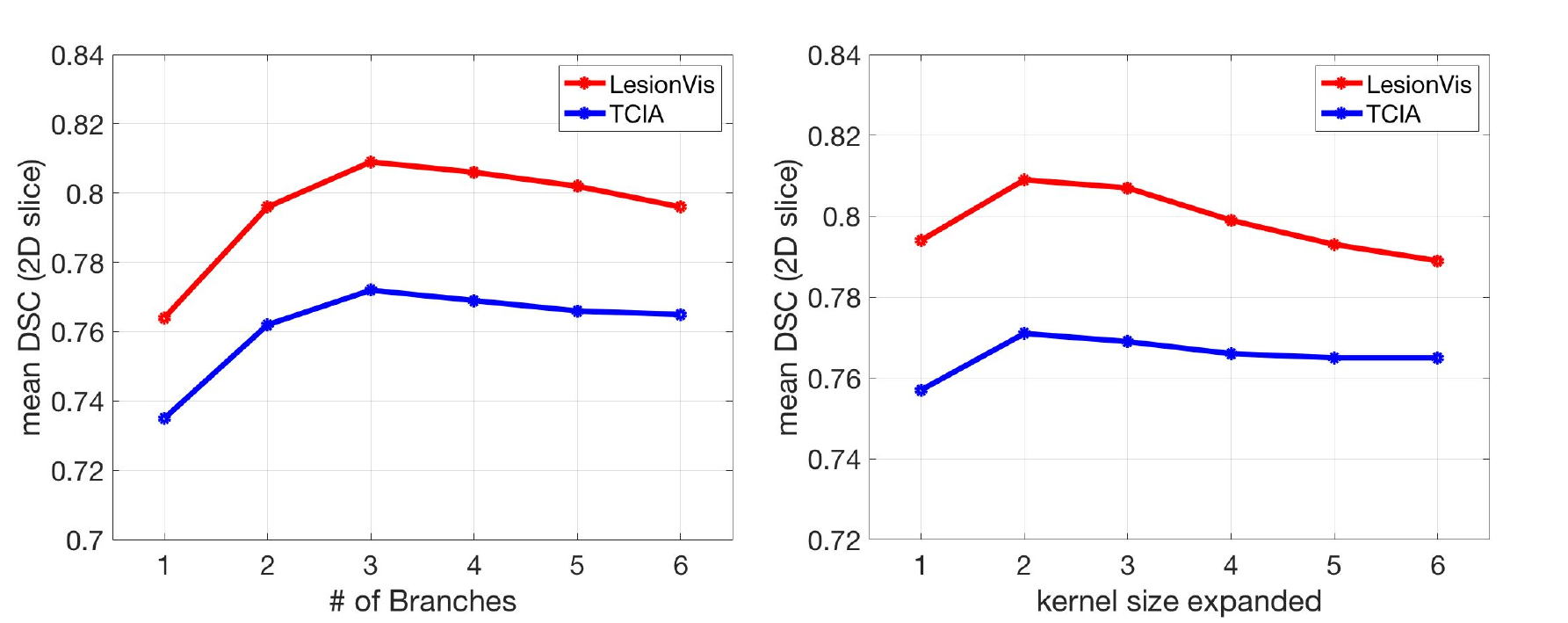}
\caption{Tuning the hyper-parameters of SiBA-Net based on LesionVis dataset. Left: Analysis on the number of scale-invariant branches. Right: Analysis on the kernel size transformation coefficient between scale-invariant branches.}
\label{fig_ana}
\end{figure}

\vspace{0.2cm}
\noindent \textbf{Further analysis:} In the design of SiBA-Net, the number of scale-invariant branches and the kernel size transformations between branches are the two most important hyper-parameters determining the performance of the network. To obtain the optimal SiBA-Net architecture, we assessed the segmentation accuracy on the RECIST-slices by changing both the number of scale-invariant branches and the coefficient for kernel size transformation between branches. The results of this analysis are shown in Figure \ref{fig_ana}. It can be seen that our model achieved the best performance when 3 branches with the coefficient for kernel size transformation set at 2 were applied. Therefore this became the default setting for the SiBA-Net in our experiments. Based on the optimal SiBA-Net architecture, we also evaluated the effect of the boundary-aware component on the segmentation accuracy. The quantitative results are shown in Table \ref{tab:t_BA}.

\begin{table} [!htb]
\centering
\caption{Effect of different structure components of SiBA-Net on the segmentation of 2D RECIST slices.}
\label{tab:t_BA}
\setlength{\tabcolsep}{7pt}

\begin{tabular}{l c c c c c}
\hline
\rule{0pt}{1.1\normalbaselineskip}
\textbf{Component}             & \multicolumn{2}{c}{Boundary-aware} & \multicolumn{2}{c}{Scale-invariant} \\ \cmidrule(r){2-3} \cmidrule(r){4-5}
\textbf{ Analysis}                         & Yes & No & Yes & No  \\ [0.1cm]
\hline
\rule{0pt}{1.1\normalbaselineskip}
Mean DSC     & $0.893$ & $0.871$ & $0.893$ & $0.891$ \\
\hline

\end{tabular}
\end{table}

\section{Discussion}
We presented a progressively-trained automatic lesion segmentation algorithm based on a customized scale-invariant and boundary-aware CNN to produce volumetric lesion segmentations by effectively leveraging 2D delineations on the RECIST-slices. We collected a large-scale dataset, named LesionVis, for the training and validation of the proposed P-SiBA approach. According to the experiments on the lung lesion cases, both quantitative and qualitative results demonstrated that the proposed approach was able to produce more accurate 3D segmentation results compared with previously proposed approaches. Several customized strategies may have contributed to this improvement. Firstly, SiBA-Net is robust when the object’s scale varies. Thus, the lesion scale variability from RECIST slices to successive slices, as well as the scale variations between time points, could be captured properly using SiBA-Net. Secondly, when the PLS approach iteratively extends the training dataset from the RECIST slices to the successive slices, the change in the data distribution can be adopted gradually and smoothly. Thirdly, the large-scale LesionVis dataset we collected is different from most of the existing databases, such as DeepLesion \cite{yan2018deep} that only contains long- and short-axis measurements of lesions on the RECIST slices. Instead, the LesionVis dataset provides accurate 2D lesion delineations on the RECIST-slice, which warrants an accurate initial segmentation for the progressive segmentation strategy. The proposed P-SiBA is a general approach that can be applied for other volumetric medical image segmentation tasks when only a limited number of 2D manual delineations are available.

\section{Conclusion}
We proposed P-SiBA as a fully automatic progressively-trained volumetric lesion segmentation approach relying on training from 2D delineations on the RECIST slices. Both quantitative and qualitative results demonstrated that P-SiBA was able to produce more accurate 3D lesion segmentations compared with previously published approaches.


{\small
\bibliographystyle{unsrt}
\bibliography{egbib}
}

\end{document}